\documentclass[letterpaper]{article}

\usepackage{graphicx}
\usepackage{amsmath,amssymb} 
\usepackage{color}
\usepackage{array}
\usepackage{multirow}
\usepackage{colortbl}
\usepackage{algorithm}
\usepackage{algorithmic}
\usepackage{fullpage}

\newcommand{\by}{\mathbf{y}}

\newcommand{\bw}{\mathbf{w}}

\newcommand{\bp}{\mathbf{p}}

\newcommand{\bo}{\mathbf{o}}

\newcommand{\be}{\mathbf{e}}

\def\be {\begin{equation}}
\def\ee {\end{equation}}
\def\beas {\begin{eqnarray*}}
\def\eeas {\end{eqnarray*}}
\def\bea {\begin{eqnarray}}
\def\eea {\end{eqnarray}}

\begin{document}

\title{Continuous Markov Random Fields for Robust Stereo Estimation} 

\author{Koichiro Yamaguchi  $\qquad$ Tamir Hazan $\qquad$  David McAllester $\qquad$ Raquel Urtasun\\
Toyota Technological Institute at Chicago
}



\maketitle

%
%

%
\begin{abstract}

In this paper we present a novel slanted-plane MRF model which reasons jointly about occlusion boundaries as well as depth. We formulate the problem as the one of
 inference in a hybrid MRF composed of both continuous (i.e., slanted 3D planes) and discrete (i.e., occlusion boundaries) random variables. This allows us to define potentials encoding  the ownership of the pixels that compose the boundary between segments, as well as potentials encoding which junctions are physically possible.
Our approach outperforms the state-of-the-art  on  Middlebury high resolution imagery \cite{Middlebury} as well as in the more challenging  KITTI dataset \cite{Geiger12}, while being more efficient than existing slanted plane MRF-based methods, taking on average 2 minutes to perform inference on high resolution imagery.


\end{abstract}

\section{Introduction}

Over the past few decades we have witnessed a great improvement in
performance of stereo algorithms. Most modern approaches frame the
problem as  inference on a Markov random field (MRF) and
utilize global optimization techniques such as graph cuts or
message passing \cite{Freeman03} to reason jointly about the depth of each
pixel in the image.

A leading approach to stereo vision uses slanted-plane MRF models which were introduced a decade ago  in \cite{BirchfieldTomasi}. 
Most methods \cite{Hong04,Bleyer05,KSK06,Deng05,Yang08,Trinh09}
assume a fixed set of superpixels on a reference image, say the left image
of the stereo pair, and model the surface under each superpixel as a slanted plane.
The MRF typically has a robust data term scoring the assigned plane in terms of
a matching score induced by the plane on the pixels contained in the superpixel.
This data term often incorporates an explicit treatment of occlusion --- pixels in one image
that have no corresponding pixel in the other image \cite{Kanade00,Kolmogorov02,Deng05,Bleyer10}.
Slanted-plane models also typically include a robust smoothness term expressing the belief that the planes assigned to
adjacent superpixels should be similar.

A major issue with slanted-plane stereo models is their computational
complexity. For example,
\cite{Bleyer10} reports an average of approximately one hour of computation for each
low-resolution Middebury stereo pair.  This makes these approaches impractical for applications such as  robotics or autonomous driving.  A main source of difficulty is that
each plane is defined by three continuous parameters and 
inference for continuous MRFs with non-convex energies is computationally challenging.

This paper contains two contributions.  First, we introduce the use of junction potentials, described below,
into this class of models.  Second, we show that particle methods can achieve strong performance with reasonable inference times
on the high-resolution, in-the-wild KITTI dataset \cite{Geiger12}.

Junction potentials originate in early line labeling algorithms
 \cite{Waltz72,Malik87}.  These algorithms assign labels to the lines of a line drawing where
the label indicate whether the line represents a discontinuity due to changes in depth (an occlusion), surface orientation  (a corner),
 lighting (a shadow) or albedo (paint).  A junction is a place where three lines meet.  Only certain
combinations of labels are physically realizable at junctions.  The constraints on label combinations at junctions
often force the labeling of the entire line drawing \cite{Waltz72}. Here, as in recent work on monocular image interpretation
\cite{Hoiem07,Saxena07a,Malik08}, we label the boundaries between image segments rather than the lines of a line drawing with labels --``left occlusion'', ``right occlusion'', ``hinge'' or
``coplanar''. In our model the occlusion labels play a role in the data term
where they are interpreted as expressing ownership
of the pixels that compose the boundary between segments --- an occlusion boundary is ``owned'' by the foreground object.


%
%

Our second contribution is to show that particle methods can be used
to implement high performance  inference in high resolution imagery with
reasonable running time.  Particle methods avoid premature commitment
to any fixed quantization of continuous variables and hence allow
a precise exploration of the continuous space.
Our particle inference method is based on
the recently developed particle convex belief propagation (PCBP)
\cite{Peng11}. We learn the contribution of each potential via  the primal-dual optimization framework of
\cite{Hazan10}.

 In the remainder of the paper we first review related work. We then introduce our continuous MRF model for stereo and show how to do learning and inference in this model.
Finally, we demonstrate the effectiveness of our approach in estimating depth from stereo pairs and show that it outperforms the state-of-the-art in the high resolution Middlebury imagery \cite{Middlebury} as well as in the more challenging KITTI dataset \cite{Geiger12}.

\newcommand{\ignore}[1]{}
 
\section{Related Work}

In the past few years much progress has been made towards solving the stereo problem, as evidenced by Scharstein et al. overview \cite{Sze02}. 
Local methods typically aggregate image statistics in a small window, thus imposing smoothness implicitly.
Optimization is usually performed using a winner-takes-all strategy, which selects for each pixel the disparity with the smallest value under some distance metric \cite{Sze02}. 
Traditional local methods \cite{Konolige97} often suffer from border bleeding effects or struggle with correspondence ambiguities. 
Approaches based on adaptive support windows \cite{Kanade94,Yoon06} adjust their computations locally to improve performance,  especially close to border discontinuities. This results in better performance at the price of more computation.

Hirschm\"uller proposed semi-global matching \cite{Hirschmueller08}, an approach which extends polynomial time 1D scan-line methods to propagate information along 16 orientations. This reduces streaking artifacts and improves accuracy compared to traditional methods. In this paper we employ this technique to compute a disparity map from which we build our potentials. 
In \cite{Cech07,Kostkova03}  disparities are `grown' from a small set of initial correspondence seeds. Though these methods produce accurate results and can be faster than global approaches, they do not provide dense matching and struggle with textureless and distorted image areas. Approaches to reduce the search space have been investigated  for global stereo methods  \cite{Wang08-1,Veksler06} as well as local methods \cite{Geiger10}.

Dense and accurate  matching can be obtained by global methods, which enforce smoothness explicitly by minimizing an MRF-based energy function.  These MRFs can be formulated at the pixel level \cite{StereoMRF}, however, the smoothness is then defined very locally. 
Slatend-plane MRF models for stereo vision were introduced in \cite{BirchfieldTomasi} and have been since very widely used \cite{Hong04,Bleyer05,KSK06,Yang08,Trinh09,Bleyer10}. 
In the context of this literature, our work has several distinctive features.
First, we use a novel model involving ``boundary labels'', ``junction potentials'', and ``edge ownership''.
Second, for inference we employ the convex form of the particle norm-product belief propagation \cite{Hazan10-ieee}, which we refer to as particle convex belief propagation (PCBP) \cite{Peng11}. In contrast, some previous works used particle belief propagation (PBP)  \cite{Koller99,PBP,Trinh09} which correspond to non-convex norm-product with the Bethe entropy approximation. The efficiency and convexity of PCBP makes it possible to evaluate our approach on hundreds of  high-resolution images \cite{Geiger12},
whereas previous empirical evaluations of slanted-plane models have largely been restricted to the low-resolution
versions of the small number of highly controlled Middlebury images.
Third, we use a training algorithm based on primal-dual approximate inference \cite{s:hazan10} which allow us to effectively learn the importance of each potential. 

\begin{figure}[t]
\vspace{-0.3cm}
\begin{center}
\begin{tabular}{cccccccc}
\includegraphics[scale=0.4]{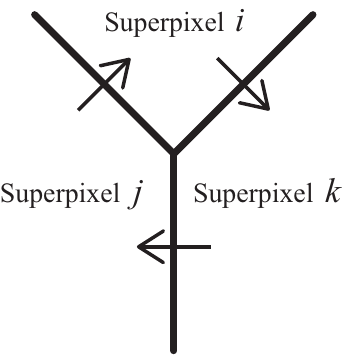}&
\includegraphics[scale=0.4]{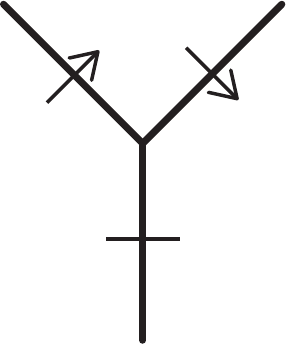}&
\includegraphics[scale=0.4]{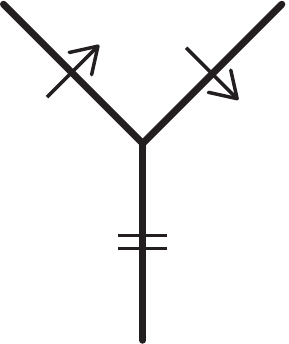}&
\includegraphics[scale=0.4]{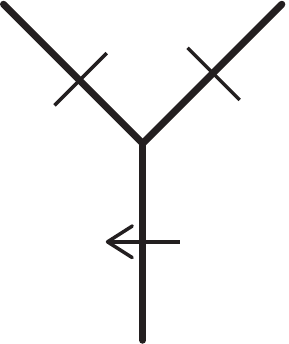}&
\includegraphics[scale=0.4]{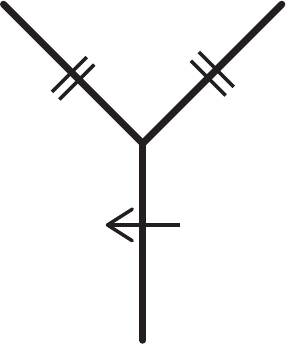}&
\includegraphics[scale=0.4]{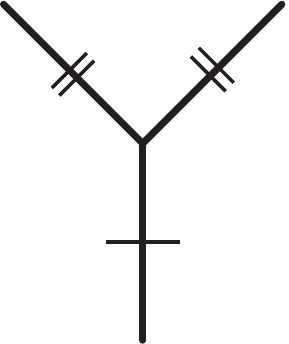}&
\includegraphics[scale=0.4]{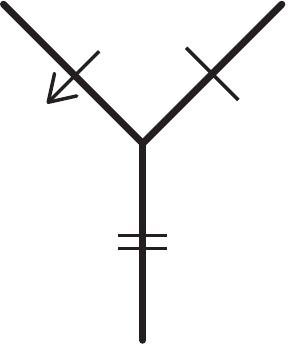}&
\hspace{3mm}
\includegraphics[scale=0.4]{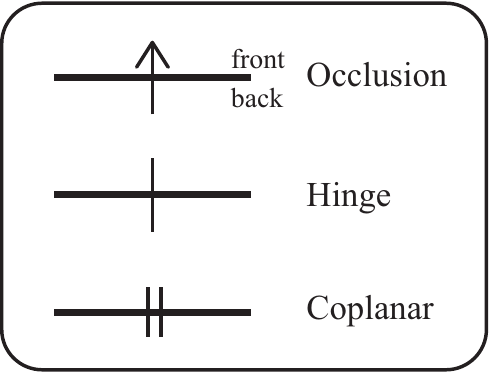}\\
(a)&(b)&(c)&(d)&(e)&(f)&(g)
\end{tabular}
\end{center}
\vspace{-0.6cm}
\caption{Impossible cases of 3-way junctions. (a) 3 cyclic occlusions, (b) hinge and 2 occlusion with opposite directions, (c) coplanar and 2 occlusion with opposite directions, (d) 2 hinge and occlusion, (e) 2 coplanar and occlusion, (f) 2 coplanar and hinge, (g) hinge, coplanar, and occlusion (superpixel with coplanar boundary is in front).}
\label{fig:3comp}
\vspace{-0.3cm}
\end{figure}

\ignore{
Throughout this work, we reason about depths of segments in stereo
images, taking into account the plane parameters of each image and the
occlusion boundaries of neighboring segments. Specifically, we
constructed an energy model, composed of continuous variables, plane
parameters, and discrete variables, occlusion boundaries and finding
the best solution using particle convex max-product \cite{Peng11}.

Energy based models are extensively used for depth estimation,
c.f. \cite{Szeliski07} and references therein. In the last decade,
the graph-cuts emerged as a successful tool for depth estimation in
energy based models. In particular, the graph-cut is the optimal
solver when there are binary labels for every pixel. However, the
graph-cuts is sub-optimal when considering many labels per pixel, and
is significantly outperformed on high-resolution images
\cite{Geiger10}. Another successful method is the max-product belief
propagation \cite{Pearl88}. However, this method optimizes a
non-convex functional, the Bethe free energy, thus in many cases get
stuck in bad local optimum. In contrast, our work uses the recent
convex max-product algorithm in \cite{Peng11} which is the only known
dual solver that is optimal when handling both discrete and continuous
labels.

Intuitively, enforcing plane consistency over an image segment enables to remove depth
estimation outliers. Outliers can occur in the initial stage of
evaluating disparities of reliable points, usually in segments without
texture, e.g. \cite{Humenberger10,Yang08,Hong04}. In other works,
outliers are removed by iteratively enforcing plane consistency in
image segments, e.g. \cite{Wei04, Bleyer12patch}. These works defer
than ours in important respects, as they perform a processing step of
plane fitting to segments, either before or after depth estimation,
while our method jointly infer disparity and plane parameters. Perhaps
the closets work to ours in this respect is \cite{Klaus06,Trinh09,Bleyer10} which jointly infer planes and disparities. \cite{Klaus06}
differs from ours in important respects, as they discretize the set of
possible planes to reason about them with max-product while our
approach reasons about continuous plane parameters with discrete
boundary parameters. \cite{Trinh09} differs from ours in important
respects, DAVID, PLEASE FILL. \cite{Bleyer10} jointly performs
segmentation and plane estimation using QPBO, thus needs to restricts
its planes through different proposal moves. In contrast, we rely on
segments derived by SLIC and UCM thus being able to infer general
planes thus able to obtain state-of-the-art results.
}

\section{Continuous MRFs for stereo}

In this section we describe our approach to joint reasoning of boundary labels and depth.  
We reason at the segment level, employing a richer representation  than a discrete disparity label. In particular, we formulate the problem as inference in a hybrid conditional random field, which contains continuous and  discrete random variables. 
The continuous random variables represent, for each segment, the disparities of all pixels contained in that segment in the form of a  3D slanted plane. The discrete random variables indicate for each pair of neighboring segments, whether they are co-planar, they form a hinge or there is a depth discontinuity (indicating which plane is in front of which).

\begin{figure}[t]
\vspace{-0.3cm}
\begin{center}
\begin{tabular}{cccccccc}
\includegraphics[scale=0.4]{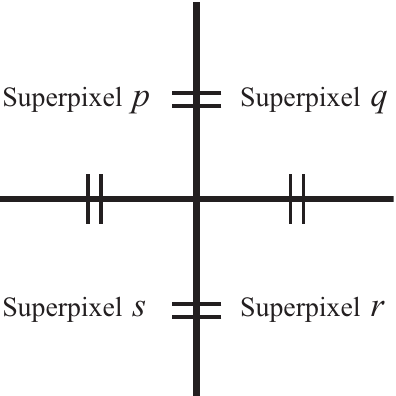}&
\includegraphics[scale=0.4]{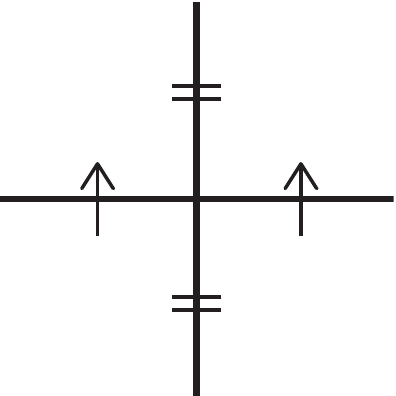}&
\includegraphics[scale=0.4]{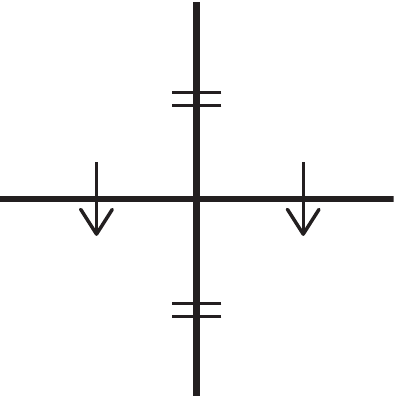}&
\includegraphics[scale=0.4]{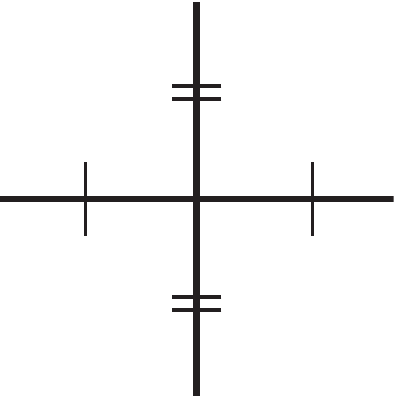}&
\includegraphics[scale=0.4]{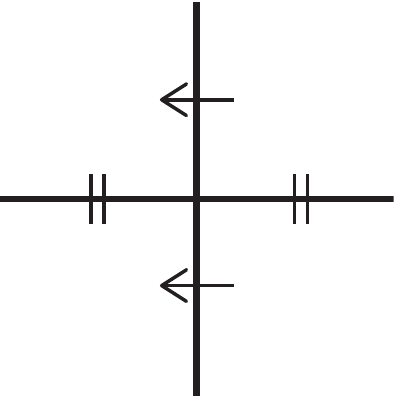}&
\includegraphics[scale=0.4]{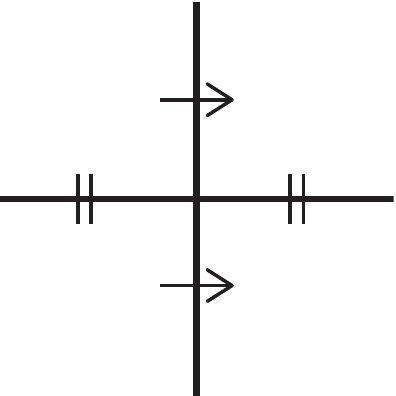}&
\includegraphics[scale=0.4]{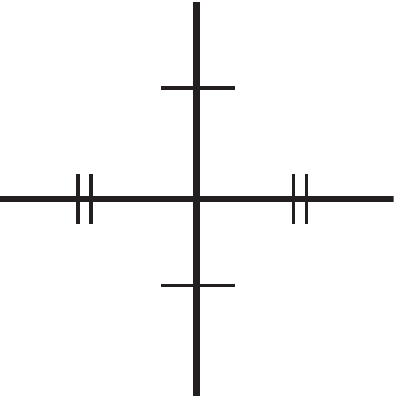}\\
(a)&(b)&(c)&(d)&(e)&(f)&(g)
\end{tabular}
\end{center}
\vspace{-0.6cm}
\caption{Valid 4-way junctions. (a) 4 coplanar boundaries, (b)-(d) 2 coplanar vertical boundaries   and 2 occlusion/hinge horizontal boundaries, (e)-(g) 2 coplanar horizontal boundaries  and 2 vertical occlusion/hinge boundariese. A 4-way junction only appears in a region of uniform color. 
}
\label{fig:4comp}
\vspace{-0.2cm}
\end{figure}

More formally, let $y_i= (\alpha_i, \beta_i, \gamma_i) \in \Re^3$ be a random variable representing the $i-$th slanted 3D plane. We can compute  the disparities of each pixel belonging to the $i-$th segment as follows 
\begin{equation}
\hat{d_i}(\mathbf{p}, \by_i) = \alpha_i(u - c_{ix}) + \beta_i(v - c_{iy}) + \gamma_i
\label{eq:disp}
\end{equation}
with $\bp = (u,v)$, and $\mathbf{c}_i = (c_{ix}, c_{iy})$ the center of the $i$-th segment.
We have defined $\gamma_i$ to be the disparity in the segment center as it improves the efficiency of PCBP inference.  
Let $o_{i,j}  \in \{co, hi, lo, ro \}$ be a discrete random variable representing whether two neighboring planes are  coplanar, form a hinge or an occlusion boundary.  Here, $lo$ implies that plane $i$ occludes plane $j$, and $ro$ represents that plane $j$ occludes plane $i$. 

We define our  hybrid conditional random field  as follows 
\[
p(\by,\bo) = \frac{1}{Z} \prod_i  \psi_i(\by_i)\prod_{\alpha}\psi_{\alpha}(\by_{\alpha})\prod_\beta\psi_{\beta}(\bo_{\beta})\prod_{\gamma} \psi_\gamma(\by_\gamma, \bo_\gamma)
\vspace{-0.2cm}
\]
where $\by$ represents the set of all 3D slanted planes, and $\bo$ the set of all discrete random variables. The unitary potentials are represented as $\psi_i$, while  $\psi_\alpha, \psi_\beta, \psi_\gamma$ encode potential functions over sets of continuos, discrete or mixture of both types of  variables.  Note that $\by$ contains three random variables for every segments in the image,  and there is a random variable $o_{i,j}$ for each pair of neighboring segments. 

In the following, we describe the different potentials we employed for our joint occlusion boundary and depth reasoning. For clarity, we describe the potentials in the log domain, i.e.,  $\bw^T\phi_i = \log(\psi_i)$ (similarly for potentials over cliques). The weights $\bw$ will be learned using structure prediction methods. 

\subsection{Occlusion Boundary and Segmentation Potentials}

\begin{figure}[t]
\begin{center}
\includegraphics[width=0.4\columnwidth]{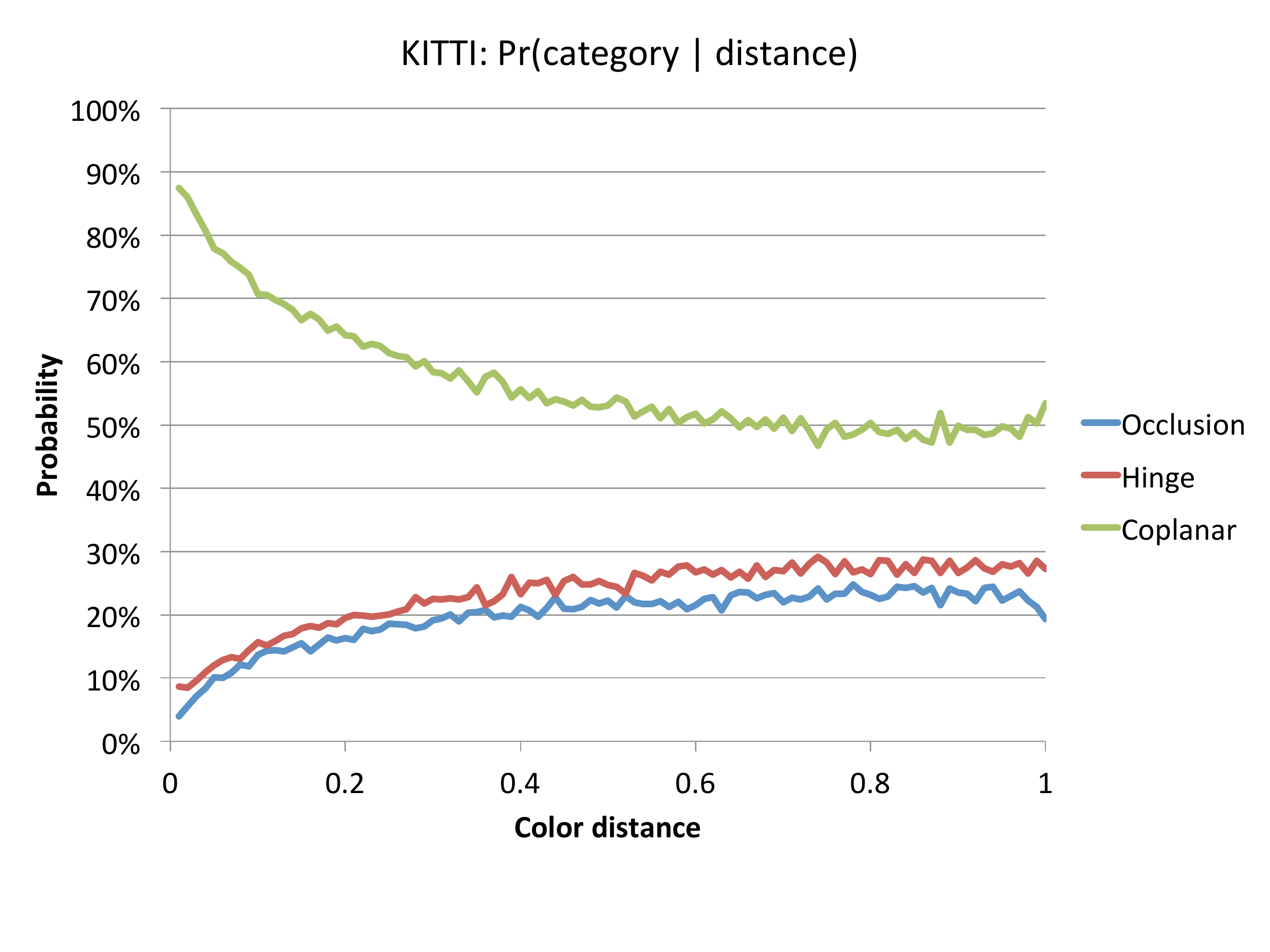}
\includegraphics[width=0.4\columnwidth]{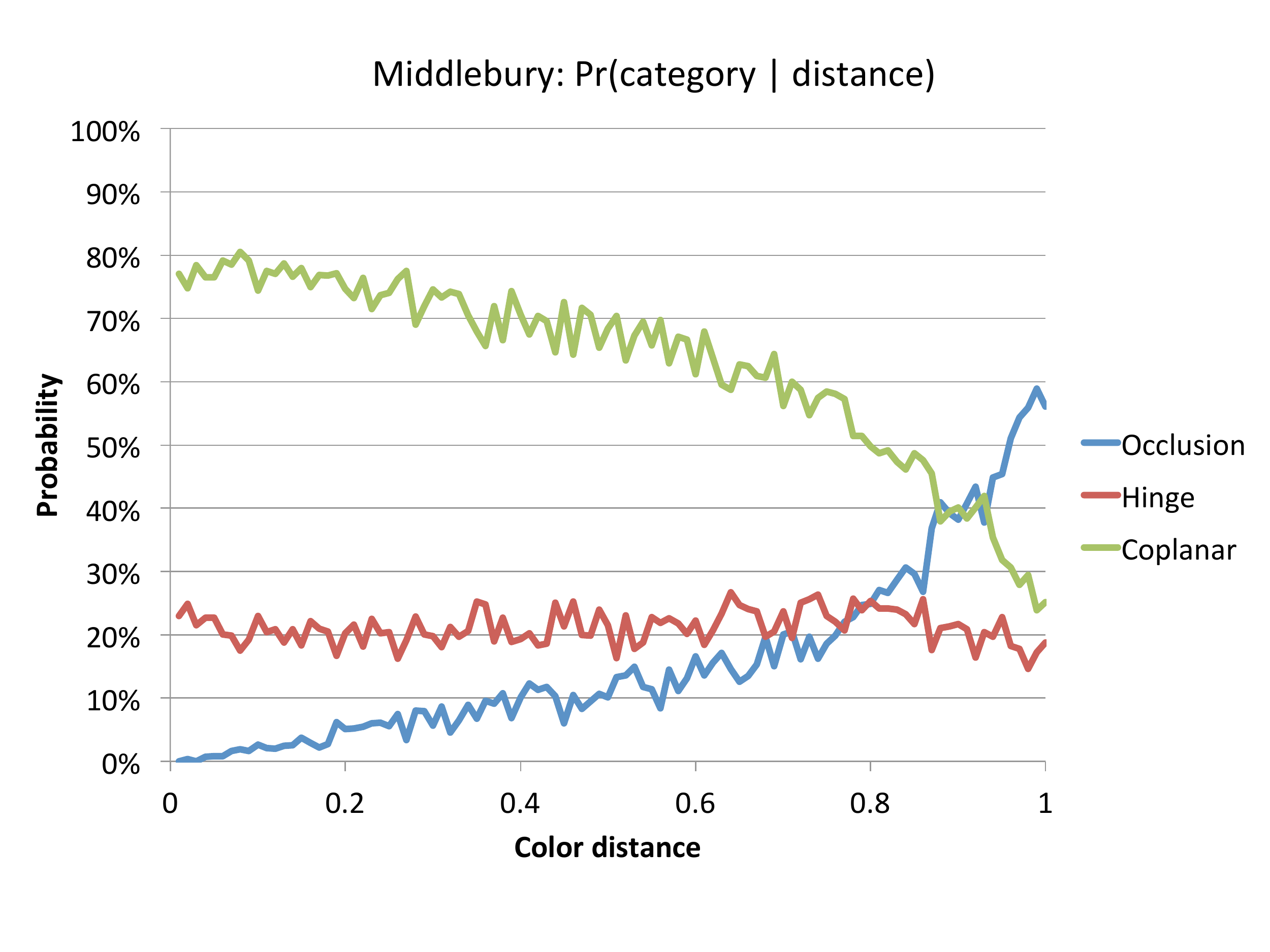}
\end{center}
\vspace{-1.1cm}
\caption{{\bf Color statistics} on (left) KITTI dataset, (right) Middlebury high-resolution}
\label{fig:color}
\end{figure}

Our approach takes as input a disparity image computed by any matching algorithm. In particular, in this paper we employ semi-global block matching \cite{Hirschmueller08}. Most matching methods return estimated disparity values on a subset of pixels.  Let  ${\cal F}$ be the set of all pixels whose initial disparity has been estimated, and let ${\cal D}(\bp)$ be the disparity of pixel $\bp \in {\cal F}$. 
Our model jointly reasons about segmentation in the form of  occlusion boundaries as well as depth. We define potentials for each of these tasks individually as well as potentials which link both tasks.
We start by  defining truncated quadratic potentials, which we will employ in the definition of some of our potentials, i.e., 
\begin{equation}
\phi^{TP}_i(\bp, \by_i, K) = \min\left(\left|\mathcal{D}(\mathbf{p}) - \hat{d_i}(\mathbf{p}, \mathbf{y}_i)\right|, K\right)^2
\end{equation}
with $K$  a constant threshold, and $\hat{d_i}(\mathbf{p}, \mathbf{y}_i)$  the disparity of pixel $\bp$ estimated as in  Eq. \ref{eq:disp}. Note that we have made the quadratic potential robust via the $\min$ function.
We now describe each of the  potentials employed  in more details.

\paragraph{{\bf Disparity potential:}} We define truncated quadratic unitary potentials for each segment expressing the fact  that the plane should agree with the results of the matching algorithm,  
\begin{equation*}
\phi_i^{\mathrm{seg}}(\mathbf{y}_i) = 
 \sum_{\mathbf{p}\in S_i \cap {\cal F}}  \phi^{TP}_i(\bp,\by_i,K)
\end{equation*}
where $S_i$ is the set of pixels in segment $i$. 

\paragraph{{\bf Boundary potential:}} We employ 3-way potentials linking our discrete and continuous variables. In particular, these potentials express the fact  that when two neighboring planes are hinge or coplanar they should agree on the boundary, and when a segment occludes another segment, the boundary should be explained by the occluder. We thus define 
\begin{equation*}
\phi_{ij}^{\mathrm{bdy1}}(o_{ij}, \by_i, \by_j) = \begin{cases}
\sum_{\mathbf{p}\in B_{ij} \cap {\cal F}} \phi_i^{TP}(\bp, \by_i,K)  & \text{if $o_{ij}$  =  $lo$ }\\
\sum_{\mathbf{p}\in B_{ij} \cap {\cal F}} \phi^{TP}_j(\bp, \by_j,K)  & \text{if $o_{ij}$  =  $ro$ }\\
\frac{1}{2}\sum_{\mathbf{p}\in B_{ij} \cap \mathcal{F}}
\phi^{TP}_i(\bp, \by_i,K)  + \phi^{TP}_j(\bp, \by_j,K) 
 & \text{if $o_{ij}=hi \vee co$}
\end{cases}
\end{equation*}
where $B_{ij}$ is the set of pixels around the boundary (within 2 pixels of the boundary) between segments $i$ and $j$.

\begin{table}[t]
\begin{center}
\begin{small}
\begin{tabular}{| c | c | c |c | c | c | c | c | c |}
\hline
& \multicolumn{2}{|c|}{  $>$ 2 pixels} & \multicolumn{2}{|c|}{  $>$ 3 pixels}& \multicolumn{2}{|c|}{  $>$ 4 pixels}& \multicolumn{2}{|c|}{  $>$ 5 pixels}\\ \hline
& {\bf Non-Occ} & {\bf Occ} & {\bf Non-Occ} & {\bf Occ}& {\bf Non-Occ} & {\bf Occ}& {\bf Non-Occ} & {\bf Occ}
\\\hline
GC+occ \cite{Kolmogorov01} & 41.92 \% & 43.64 \% & 34.04 \% & 35.85 \%& 29.79 \% & 31.59 \% & 27.05 \% & 28.82 \%
\\\hline
OCV-BM \cite{Bradski00} & 29.18 \% & 31.65 \%  & 24.94 \% & 27.32 \%& 23.21 \% & 25.48 \% & 22.02 \% & 24.18 \%
 \\\hline
CostFilter \cite{Rhemann11} & 28.46 \% & 29.96 \%   & 20.74 \% & 22.13 \%& 17.24 \% & 18.47 \% & 15.40 \% & 16.50 \%
\\\hline
GCS \cite{Cech07} & 21.87 \% & 23.76 \%  & 14.21 \% & 15.92 \% & 10.61 \% & 12.12 \% & 8.55 \% & 9.90 \%
\\\hline
GCSF \cite{Cech11} & 20.75 \% & 22.69 \%   & 13.02 \% & 14.77 \% & 9.48 \% & 11.02 \% & 7.48 \% & 8.84 \%
 \\\hline
SDM \cite{Kostkova03} & 19.01 \% & 20.89 \%   & 11.95 \% & 13.65 \% & 8.98 \% & 10.47 \% & 7.35 \% & 8.66 \%
\\ \hline
ELAS \cite{Geiger10} & 11.25 \% & 13.71 \%  & 7.60 \% & 9.77 \% & 5.91 \% & 7.79 \% & 4.87 \% & 6.52 \%
\\\hline
OCV-SGBM \cite{Hirschmueller08} & 12.48 \% & 14.86 \% & 7.40 \% & 9.54 \% & 5.51 \% & 7.43 \% & 4.52 \% & 6.21 \%
\\ \hline 
{\bf Ours} & {\bf 9.03} \% & {\bf 11.58} \% & {\bf 4.47} \% & {\bf 6.66} \%  & {\bf 3.13} \% & {\bf 5.05} \% & {\bf 2.51} \% & {\bf 4.23} \%
\\\hline
\end{tabular}
\end{small}
\end{center}
\vspace{-0.2cm}
\caption{Comparison with the state of the art KITTI dataset}
\label{tab:kitti}
\end{table}

\begin{figure}[t]
\begin{center}

%
%
%

\vspace{0mm}
\includegraphics[width=4.5cm]{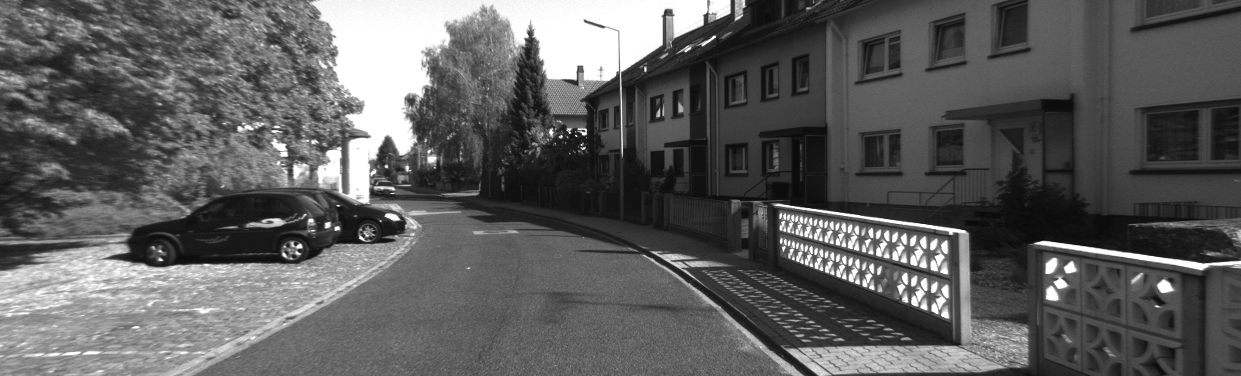}
\includegraphics[width=4.5cm]{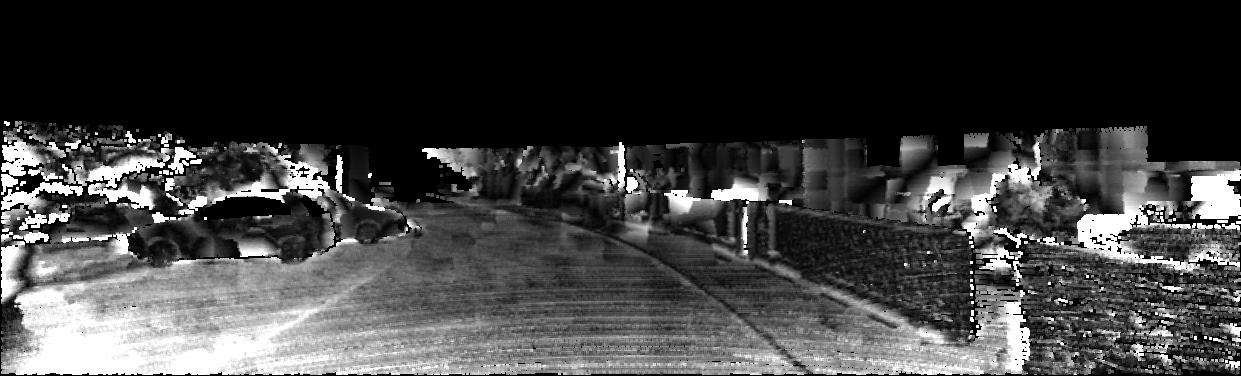}

\includegraphics[width=4.5cm]{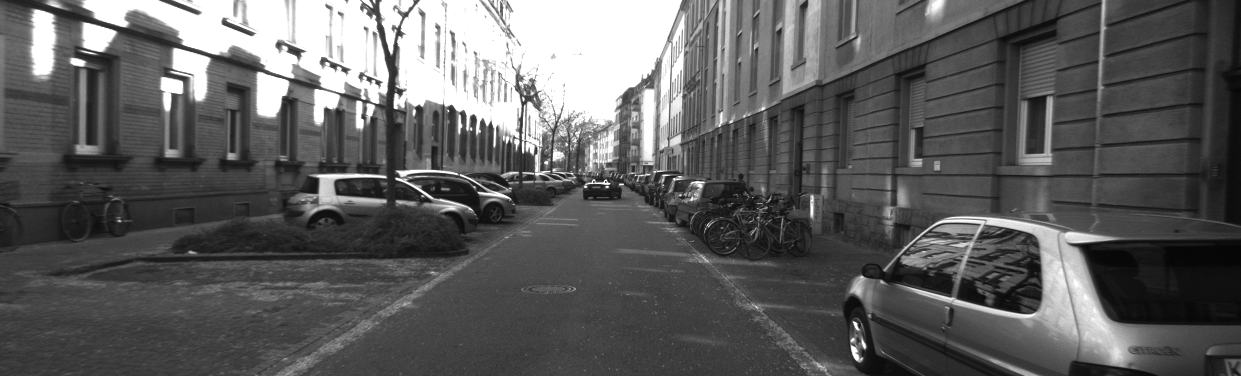}
\includegraphics[width=4.5cm]{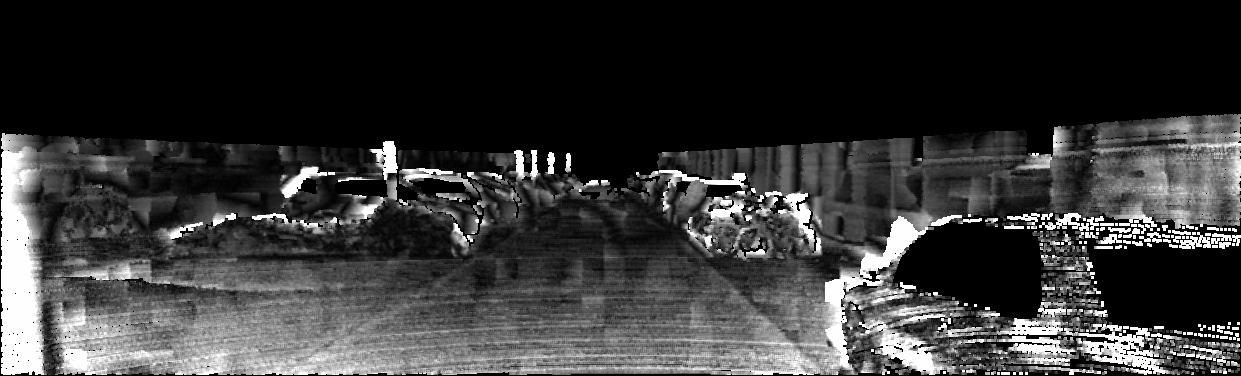}
\end{center}
\vspace{-0.7cm}
\caption{Examples from the KITTI. (Left) Original images. (Right) Disparity errors.}
\label{fig:example}
\vspace{-0.2cm}
\end{figure}

\paragraph{{\bf Compatibility potential:}} We introduce an additional potential which ensures that the discrete occlusion labels match well the disparity  observations.  
We do so by defining  $\phi^{\mathrm{occ}}_{ij}(\by_{\mathrm{front}},\by_{\mathrm{back}})$ to be a penalty term which penalizes occlusion boundaries that are not supported by the data 
\begin{equation*}
\phi^{\mathrm{occ}}_{ij}(\by_{\mathrm{front}}, \by_{\mathrm{back}}) = \begin{cases}
\lambda_{\mathrm{imp}} &
 \text{if $\exists {\mathbf{p}\in B_{ij}}: \hspace{0.2cm}
 \hat{d_i}(\mathbf{p},\by_{\mathrm{front}}) < \hat{d_j}(\mathbf{p}, \by_{\mathrm{back}})$ } \\
0 & \text{otherwise}
\end{cases}
\end{equation*}
We also define  $\phi^{\mathrm{neg}}_{ij}(\by_i)$ to be a function which penalizes negative disparities 
\begin{equation*}
\phi^{\mathrm{neg}}_{ij}(\by_i) = \begin{cases}
\lambda_{imp} & \text{if $\min_{\mathbf{p}\in B_{ij}} \hat{d_i}(\mathbf{p}, \by_i) < 0$} \\
0 & \text{otherwise}
\end{cases}
\end{equation*}
We impose a regularization on the type of occlusion boundary, where we prefer simpler explanations (i.e., coplanar is preferable than hinge which is more desirable than occlusion). We encode this preference by defining  $\lambda_{\mathrm{occ}} >\lambda_{\mathrm{hinge}}> 0$.
We thus define our computability potential 
\begin{equation*}
\begin{small}
\phi_{ij}^{\mathrm{bdy2}}(o_{ij}, \by_i, \by_j) = \begin{cases}
\lambda_{\mathrm{occ}} + \phi^{\mathrm{neg}}_{ij}(\by_i) + \phi^{\mathrm{neg}}_{ij}(\by_j) + \phi^{\mathrm{occ}}_{ij}(\by_i, \by_j)  & \text{if $o_{ij}=lo$} \\
\lambda_{\mathrm{occ}}  + \phi^{\mathrm{neg}}_{ij}(\by_i) + \phi^{\mathrm{neg}}_{ij}(\by_j) + \phi^{\mathrm{occ}}_{ij}(\by_j, \by_i)& \text{if $o_{ij}=ro$} \\
\lambda_{\mathrm{hinge}} +  \phi^{\mathrm{neg}}_{ij}(\by_i) + \phi^{\mathrm{neg}}_{ij}(\by_j) + \frac{1}{|B_{ij}|}\sum_{\mathbf{p}\in B_{ij}} \Delta d_{ij}    & \text{if $o_{ij}= hi$} \\
\phi^{\mathrm{neg}}_{ij}(\by_i) + \phi^{\mathrm{neg}}_{ij}(\by_j)  + \frac{1}{|S_i \cup S_j|}\sum_{\mathbf{p}\in S_i \cup S_j}\Delta d_{i,j}  & \text{if $o_{ij}=co$}
\end{cases}
\end{small}
\end{equation*}
with  $\Delta d_{i,j} =(\hat{d_i}(\mathbf{p}, \by_i) - \hat{d_j}(\mathbf{p}, \by_j))^2$. 

\begin{table}[t]
\begin{center}
\begin{scriptsize}
\begin{tabular}{| c | c | c | c | c | c | c | c | c | c | c |}
\hline
 & \multicolumn{2}{|c|}{  $>$ 1 pixel} & \multicolumn{2}{|c|}{  $>$ 2 pixels} & \multicolumn{2}{|c|}{  $>$ 3 pixels} & \multicolumn{2}{|c|}{  $>$ 4 pixels} & \multicolumn{2}{|c|}{  $>$ 5 pixels} \\
 \cline{2-11}
 & {\bf N.-Occ} & {\bf Occ} & {\bf N.-Occ} & {\bf Occ} & {\bf N.-Occ} & {\bf Occ} & {\bf N.-Occ} & {\bf Occ} & {\bf N.-Occ} & {\bf Occ} \\
 \hline 
 GC+occ \cite{Kolmogorov01} & 23.8 \% &- & 16.6 \% & - & 13.9 \% & - & 12.5 \% & - & 11.5 \% & - \\
 \hline
 EBP \cite{Pedro06} & 14.3 \% & - & 10.3 \% & - & 9.4 \% & - & 9.0 \% & - & 8.7 \% & - \\
 \hline
 GCS \cite{Cech07} & 13.2 \% & - & 9.0 \% & -  & 7.4 \% & - & 6.5 \% & - & 5.9 \% & - \\
 \hline
 SDM \cite{Kostkova03} & 12.8 \% & - & 9.3 \% & - & 8.2 \% & - & 7.7 \% & - & 7.3 \% & - \\
 \hline
 ELAS \cite{Geiger10} & 7.1 \% & 23.4 \% & 4.7 \% & 17.2 \% & 3.9 \% & 14.3 \% & 3.5 \% & 12.8 \% & 3.2 \% & 11.8 \% \\
 \hline
 OCV-SGBM \cite{Hirschmueller08} & 7.0 \% & 24.4 \% & 5.9 \% & 22.9 \% & 5.5 \% & 22.4 \% & 5.3 \% & 22.1 \% & 5.2 \% & 21.8 \% \\
 \hline
 \textbf{Ours} & {\bf 5.1} \% & {\bf 17.5} \% & {\bf 3.3} \% & {\bf 14.3} \% & {\bf 2.8} \% & {\bf 13.0} \% & {\bf 2.5} \% & {\bf 12.2} \% & {\bf 2.3} \% & {\bf 11.6} \% \\
 \hline
\end{tabular}
\end{scriptsize}
\end{center}
\caption{Comparison with the state-of-the-art on  Middlebury high-resolution imagery. The baselines are provided by the author of \cite{Geiger10}}
\label{tab:midd}
\end{table}

\begin{table}[t]
\begin{center}
\begin{small}
\begin{tabular}{| c  | c | c | c | c | c | c | c | c | c |}
\hline
 & Super- &  \multicolumn{2}{|c|}{  $>$ 2 pixels} & \multicolumn{2}{|c|}{  $>$ 3 pixels}& \multicolumn{2}{|c|}{  $>$ 4 pixels}& \multicolumn{2}{|c|}{  $>$ 5 pixels}\\ \cline{3-10}
 & pixels  & {\bf N.-Occ} & {\bf Occ} & {\bf N.-Occ} & {\bf Occ}& {\bf N.-Occ} & {\bf Occ}& {\bf N.-Occ} & {\bf Occ}
\\ \hline 
UCM & 91.7 &  58.06\% & 58.85\% & 52.13\% & 52.86\% & 48.67\% & 49.35\% & 46.26\% & 46.88\% \\ \hline
SLIC & 978.6 & 9.21\% & 11.73\% & 4.60\% & 6.75\% & 3.26\% & 5.15\% & 2.63\% & 4.32\% \\ \hline
SLIC & 1198.8 &  9.16\% & 11.76\% & 4.62\% & 6.86\% & 3.26\% & 5.23\% & 2.62\% & 4.39\% \\ \hline
UCM+SLIC & 1191.7  & 9.03\% & 11.58\% & 4.47\% & 6.66\%  & 3.13\% & 5.05\% & 2.51\% & 4.23\% \\
\hline
\end{tabular}
\end{small}
\end{center}
\caption{Difference in performance when employing different segmentation methods to compute superpixels on the KITTI dataset. Note that employing an intersection of SLIC+UCM superpixels works best for the same amount of superpixels. }
\label{tab:ucm_kitti}
\end{table}

\paragraph{{\bf Junction Feasibility: }} Following work on occlusion boundary reasoning \cite{Malik87,Hoiem07}, we utilize higher order potentials to encode whether a junction of three planes is possible. We refer the reader to Fig. \ref{fig:3comp} for an illustration of these cases.  We thus define the compatibility of a junction $\{i,j,k\}$ to be 
\begin{equation*}
\phi_{ijk}^{\mathrm{jct}}(o_{ij}, o_{jk}, o_{ik}) = \begin{cases}
\lambda_{\mathrm{imp}} & \mathrm{if \ impossible \ case} \\
0 & \mathrm{otherwise}
\end{cases}
\end{equation*}
We also defined a potential encoding the feasibility of a junction of four planes (see Fig. \ref{fig:4comp}) as follows 
\begin{equation*}
\phi_{pqrs}^{crs}(o_{pq}, o_{qr}, o_{rs}, o_{ps}) = \begin{cases}
\lambda_{\mathrm{imp}} & \mathrm{if \ impossible \ case} \\
0 & \mathrm{otherwise}
\end{cases}
\end{equation*}
Note that, although these potentials are high order, they only involve variables with small number of states, i.e., 4 states.

\paragraph{{\bf Potential for color similarity: }} Finally, we employ a simple color potential to reason about segmentation, which is defined in terms of the $\chi$-squared distance between color histograms of neighboring segments. 
This potential encodes the fact that we expect segments which are coplanar to have similar color statistics (i.e., histograms), while the entropy of this distribution is higher when the planes form an  occlusion boundary or a hinge. Note that this trend is shown in Fig. \ref{fig:color} (left) for the KITTI \cite{Geiger12} dataset. 
The statistics are less meaningful in the case of the Middelbury high resolution imagery \cite{Middlebury}, as this dataset is captured in a control environment.
We thus reflect these statistics in the following potential
\begin{equation*}
\phi_{ij}^{\mathrm{col}}(o_{ij}) = \begin{cases}
\min\left(\kappa \cdot \chi^2(h_i, h_j), \lambda_{\mathrm{col}}\right) & \text{if $o_{ij}=co$}\\
\lambda_{\mathrm{col}} & \text{otherwise}
\end{cases}
\end{equation*}
with $\kappa$ a scalar and $\chi^2(h_i, h_j)$ the $\chi$-squared distance between the color histograms of segments $i$ and $j$.

\subsection{Inference in Continuous MRFs}

Now that  we have defined the model, we can turn our attention to inference, which is defined as 
 computing the MAP estimate as follows 
\begin{eqnarray}
(\text{inference})\hspace{0.3cm} \, \,\text{arg} \max_{\by,\bo} \frac{1}{Z} \prod_i  \psi_i(\by_i)\prod_{\alpha}\psi_{\alpha}(\by_{\alpha})\prod_\beta\psi_{\beta}(\bo_{\beta})\prod_{\gamma} \psi_\gamma(\by_\gamma, \bo_\gamma)
\vspace{-0.2cm}
\label{eq:map}
\end{eqnarray}
Inference in this model is in general NP hard. Our inference is also particularly challenging since, unlike traditional MRF stereo formulations, we have defined a hybrid MRF, which reasons about continuous as well as discrete variables.

While there is a vast literature on discrete MRF inference, only a few attempts have focussed on  solving the continuous case. 
The exact MAP solution can only be recovered in very restrictive cases. For example when the potentials are quadratic and diagonally dominated, an algorithm called  Gaussian Belief propagation \cite{GBP} returns the optimal solution. 
For general potentials, one can approximate the messages using mixture models, or via particles. 
In this paper we make use of  particle convex belief propagation (PCBP)~\cite{Peng11}, a technique that  is guarantee to converge and gradually approach the optimum. This works very well in practice, yielding   state-of-the-art results. 

PCBP is an iterative algorithm that works as follows: For each random variable, particles are sampled around the current solution. These samples act as labels in a discretized MRF which is solved to convergence using convex belief propagation~\cite{Hazan10-ieee}. The current solution is then updated with the MAP estimate obtained on the discretized MRF. This process is repeated for a fixed number of iterations. In our implementation, we use the distributed message passing algorithm of~\cite{Schwing11} 
to solve the discretized MRF at each iteration. 
Algorithm~\ref{algo:pbp} depicts PCBP for our formulation. 
At each iteration, to balance the trade off between exploration and exploitation, we decrease the values of the standard deviations $\sigma_\alpha, \sigma_\beta$ and $\sigma_\gamma$ of the normal distributions from which the plane random variables are drawn.

\vspace{-0.2cm}
\begin{algorithm}[h]
  \begin{algorithmic} 
  \caption{PCBP for stereo estimation and occlusion boundary reasoning} 	
\label{algo:pbp}
\STATE Set $N$
\STATE Initialize slanted planes $\by_i^0 = (\alpha_i^0, \beta_i^0,\gamma_i^0)$ via local fitting $\forall i$
\STATE Initialize $\sigma_\alpha, \sigma_\beta$ and $\sigma_\gamma$
\FOR{$t = 1$ to $\#$iters}
\STATE Sample $N$ times $\forall i$ from  $\alpha_i \sim {\cal N}(\alpha_i^{t-1},\sigma_\alpha)$, $\beta_i \sim {\cal N}(\beta_i^{t-1},\sigma_\beta)$, $\gamma_i \sim {\cal N}(\gamma_i^{t-1},\sigma_\gamma)$
\STATE $(\bo^{t},\by^t)$ $\leftarrow $ Solve the discretized MRF using convex BP
\STATE Update  $\sigma^c_{\alpha} = \sigma^c_{\beta} = 0.5\times\exp(-c/10)$ and $\sigma^c_{\gamma} = 5.0\times\exp(-c/10)$
\ENDFOR
\STATE Return $\bo^t$, $\by^t$
  \end{algorithmic} 
\end{algorithm}
\vspace{-0.15cm}

\subsection{Learning in Continuous MRFs}

We employ the  algorithm of \cite{Hazan10} for learning. 
Given a set of training images and corresponding depth labels, the goal of learning is to estimate the weights which minimize the surrogate partition loss. 
However, our learning problem, as opposed to the one defined in \cite{Hazan10}, contains a mixture  of continuous and discrete variables.  Therefore the surrogate partition loss in our setting requires to integrate over the continuous variables. 
We note that our continuous variables have robust quadratic potentials, thus integrating over them can be efficiently estimated by discretizing the continuous variables. In practice, summing over $30$ particles gives a good approximation for the integral.  

 \begin{table}[t]
\begin{center}
\begin{scriptsize}
\begin{tabular}{| c | c | c | c | c | c | c | c | c | c | c | c |}
\hline
 & Super- & \multicolumn{2}{|c|}{  $>$ 1 pixel} & \multicolumn{2}{|c|}{  $>$ 2 pixels} & \multicolumn{2}{|c|}{  $>$ 3 pixels} & \multicolumn{2}{|c|}{  $>$ 4 pixels} & \multicolumn{2}{|c|}{  $>$ 5 pixels} \\
 \cline{3-12}
 & pixels & {\bf N.-Occ} & {\bf Occ} & {\bf N.-Occ} & {\bf Occ} & {\bf N.-Occ} & {\bf Occ} & {\bf N.-Occ} & {\bf Occ} & {\bf N.-Occ} & {\bf Occ} \\
 \hline 
UCM  & 259.0 & 35.5\% & 48.2\% & 26.7\% & 37.8\% & 21.3\% & 31.4\% & 17.7\% & 27.1\% & 15.2\% & 23.7\% \\
\hline
SLIC & 1787.6 & 5.4\% & 17.8\% & 3.6\% & 14.4\% & 3.0\% & 13.1\% & 2.8\% & 12.3\% & 2.6\% & 11.7\% \\
\hline
SLIC & 2066.1 & 5.3\% & 17.8\% & 3.5\% & 14.5\% & 3.0\% & {\bf 13.0}\% & 2.7\% & {\bf 12.2}\% & 2.5\% & {\bf 11.6}\% \\
\hline
UCM+SLIC & 2042.6 & {\bf 5.1}\% & {\bf 17.5}\% & {\bf 3.3}\% & {\bf 14.3}\% & {\bf 2.8}\% & {\bf 13.0}\% & {\bf 2.5}\% & {\bf 12.2}\% & {\bf 2.3}\% & {\bf 11.6}\% \\
\hline
\end{tabular}
\end{scriptsize}
\end{center}
\caption{Performance changes when employing different segmentation methods to compute superpixels on the Middlebury high-resolution dataset. Employing an intersection of SLIC+UCM superpixels works best for the same amount of superpixels.}
\label{tab:ucm_midd}
\end{table}

\begin{figure}[t]
\begin{center}
\includegraphics[width=0.32\columnwidth]{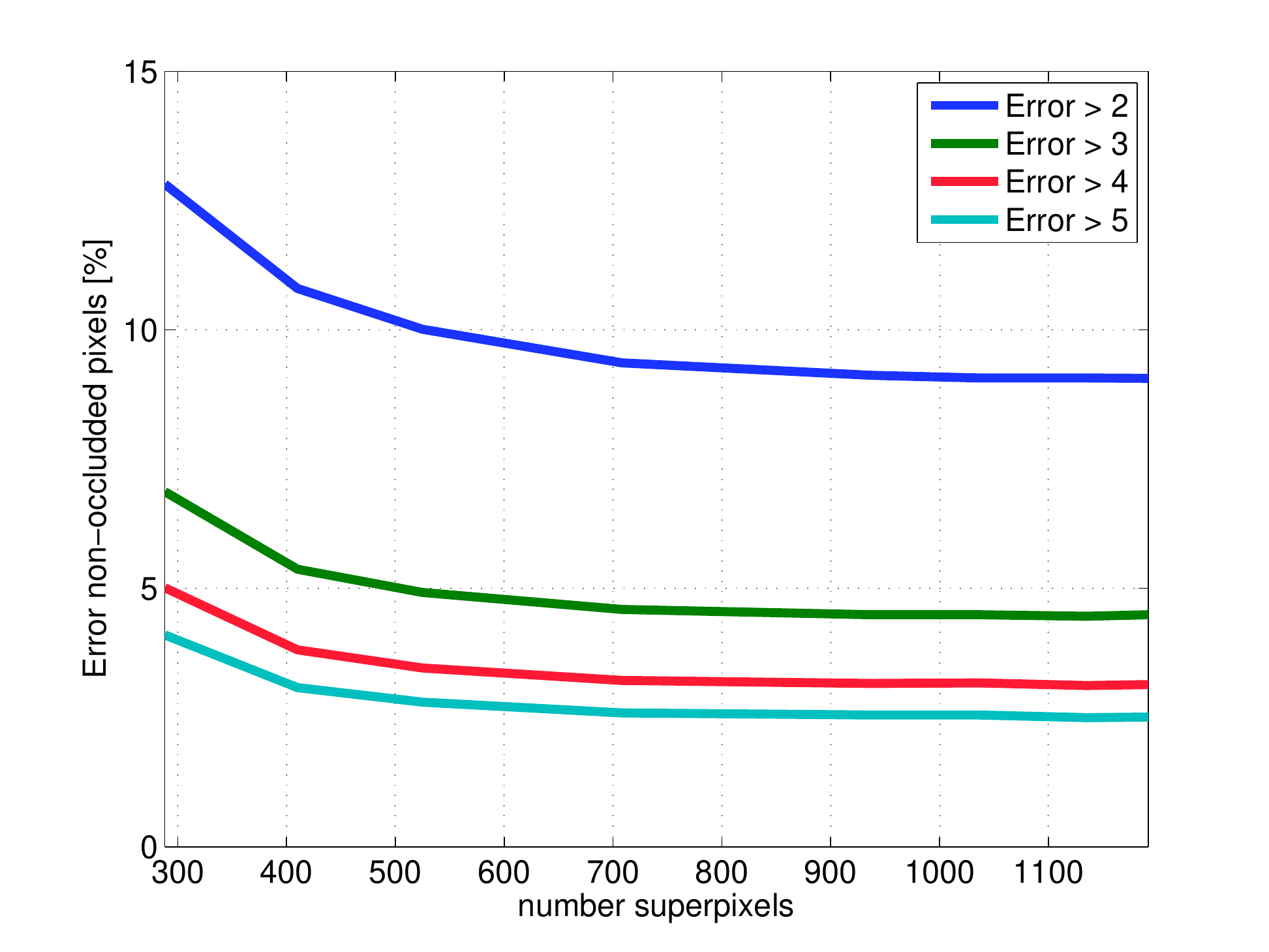}\hspace{-0.5cm}
\includegraphics[width=0.32\columnwidth]{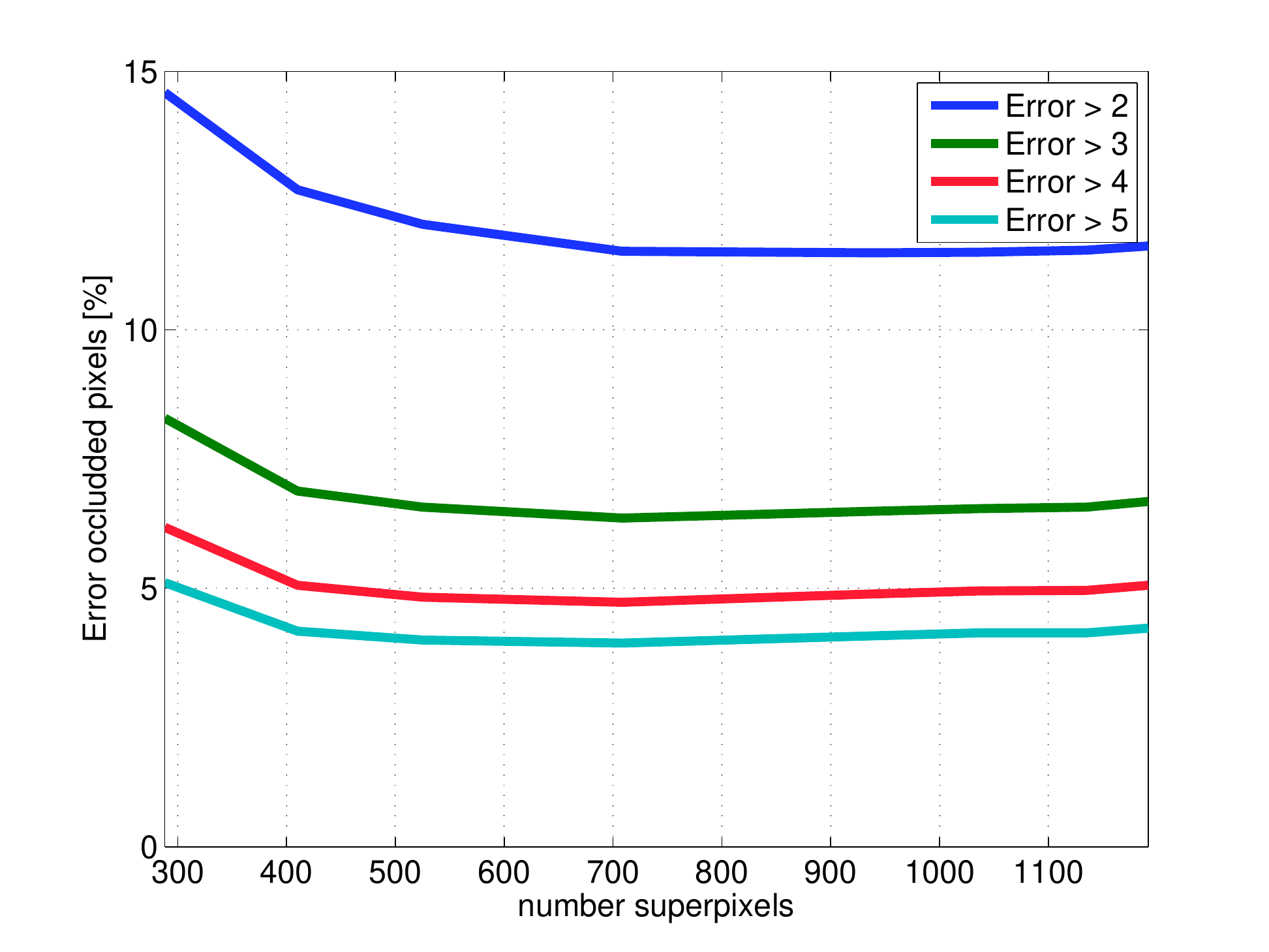}\hspace{-0.5cm}
\includegraphics[width=0.32\columnwidth]{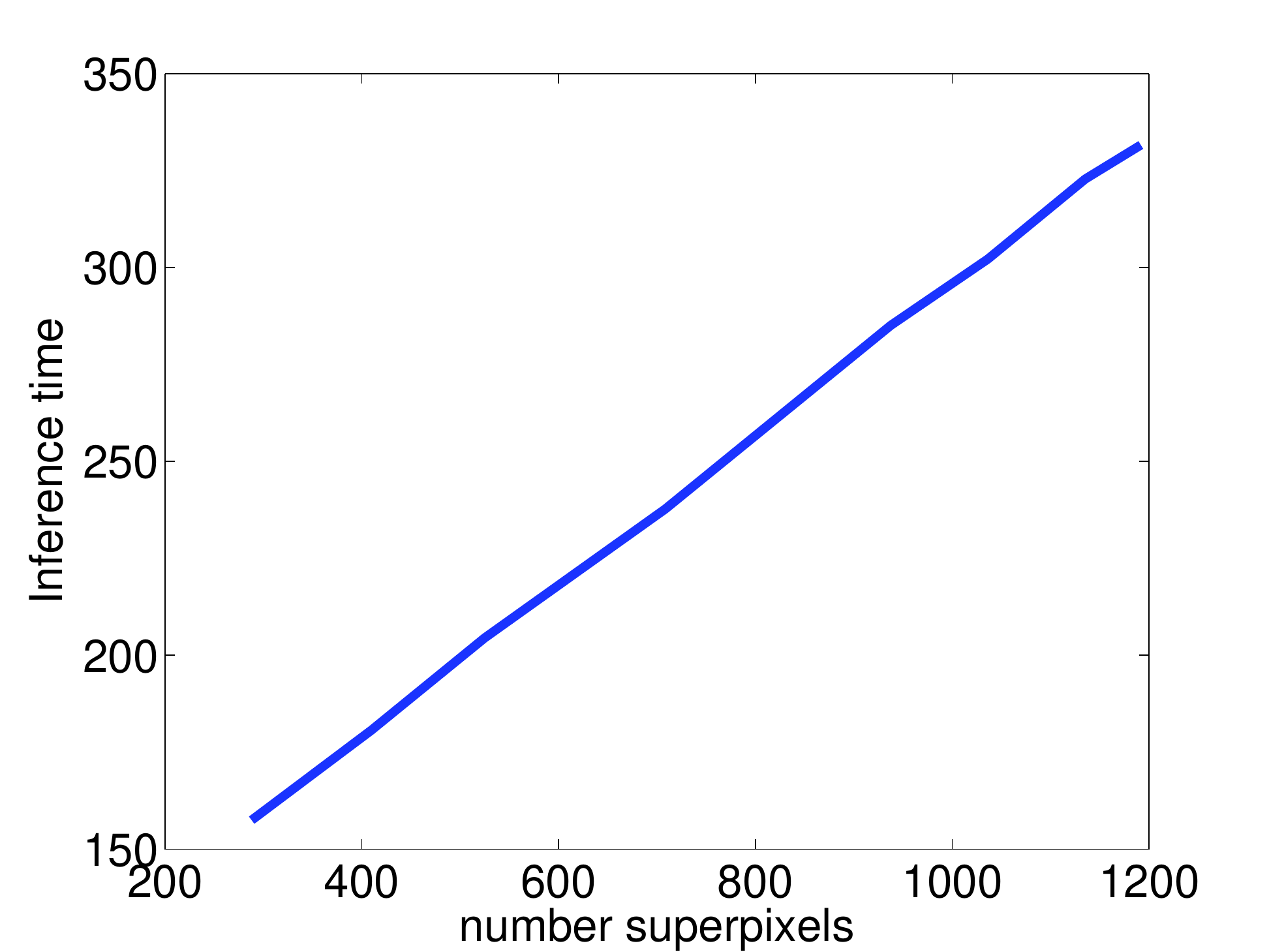}
\end{center}
\vspace{-0.6cm}
\caption{{\bf Importance of the number of superpixels:} KITTI results as a function of the number of superpixels. Even with a small number our approach still outperforms the baselines. (Right) The inference time scales linearly with the number of superpixels}
\label{fig:time}
\end{figure}

\section{Experimental Evaluation}

We perform exhaustive experiments on two publicly available datasets: Middebury high resolution images \cite{Middlebury} as well as the more challenging KITTI dataset \cite{Geiger12}. 
For all experiments, we employ the same parameters which have been validated on the training set. We use a disparity difference threshold $K = 5.0$ pixels, and set  $\lambda_{\mathrm{occ}} = 15$, $\lambda_{\mathrm{hinge}} = 3$, $\lambda_{\mathrm{imp}} = 30$
 and  $\lambda_{\mathrm{col}} = 30$. For the color potential, we use a color histogram with 64 bins and set $\kappa = 60$. Unless otherwise stated, we employ 10 particles and 5 iterations of re-sampling for PCBP \cite{Peng11}, and run each iteration of convex BP  to convergence. 
 For learning, we use a value of $C$ equal to the number of examples and unless otherwise stated use a CRF, i.e., $\epsilon=1$. 
 We learned the importance of each potential, thus $6$ parameters. 
We employ two different metrics. The first one measures  the average number of non-occluded pixels which error is bigger than  a fixed threshold.
To test the extrapolation capabilities of the different approaches, the second metric computes the average number of pixels (including the occluding ones) which error is bigger than  a fixed threshold.

\begin{figure}[t]
\begin{center}
\includegraphics[width=0.35\columnwidth]{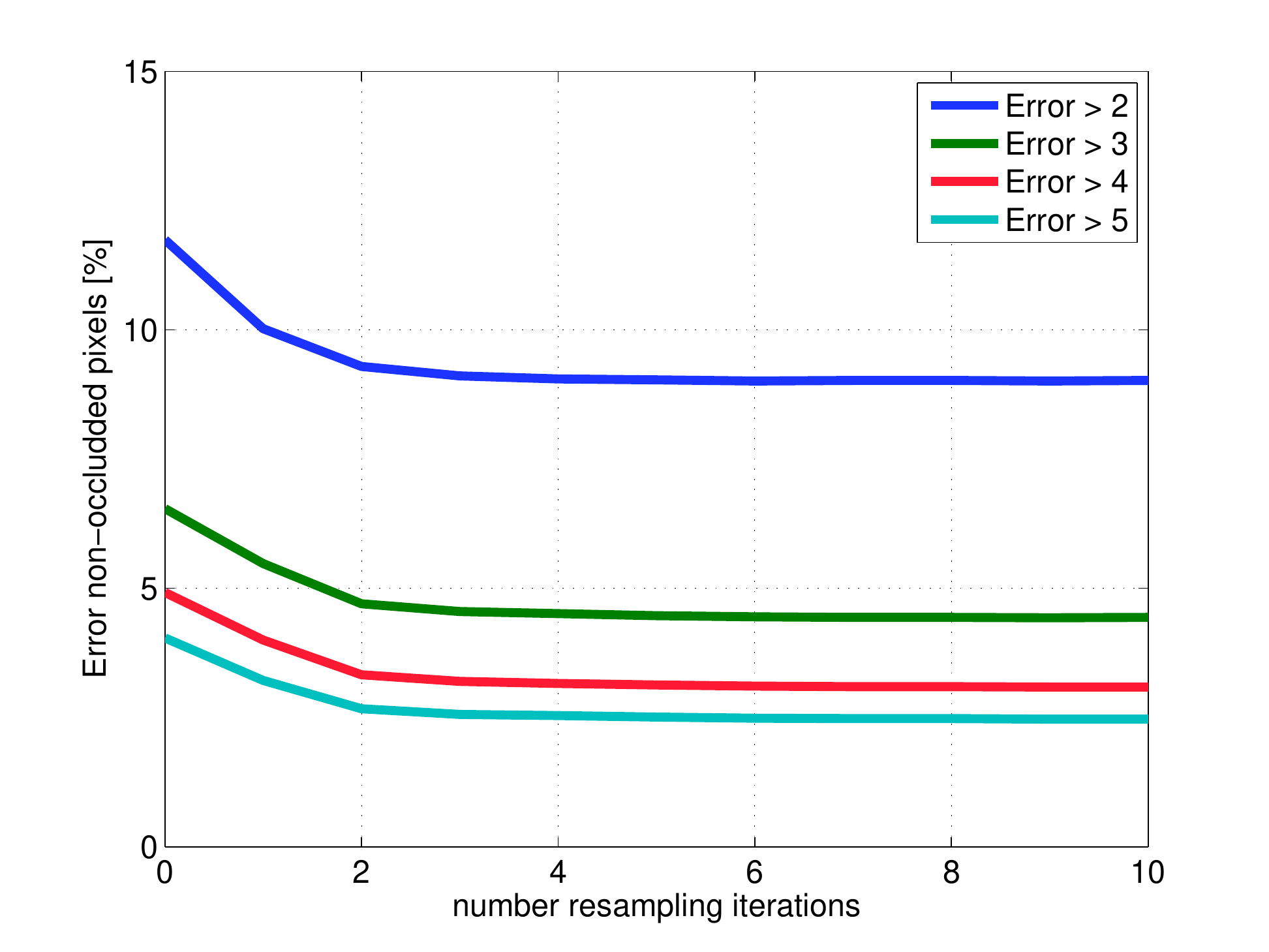}\hspace{0.2cm}
\includegraphics[width=0.35\columnwidth]{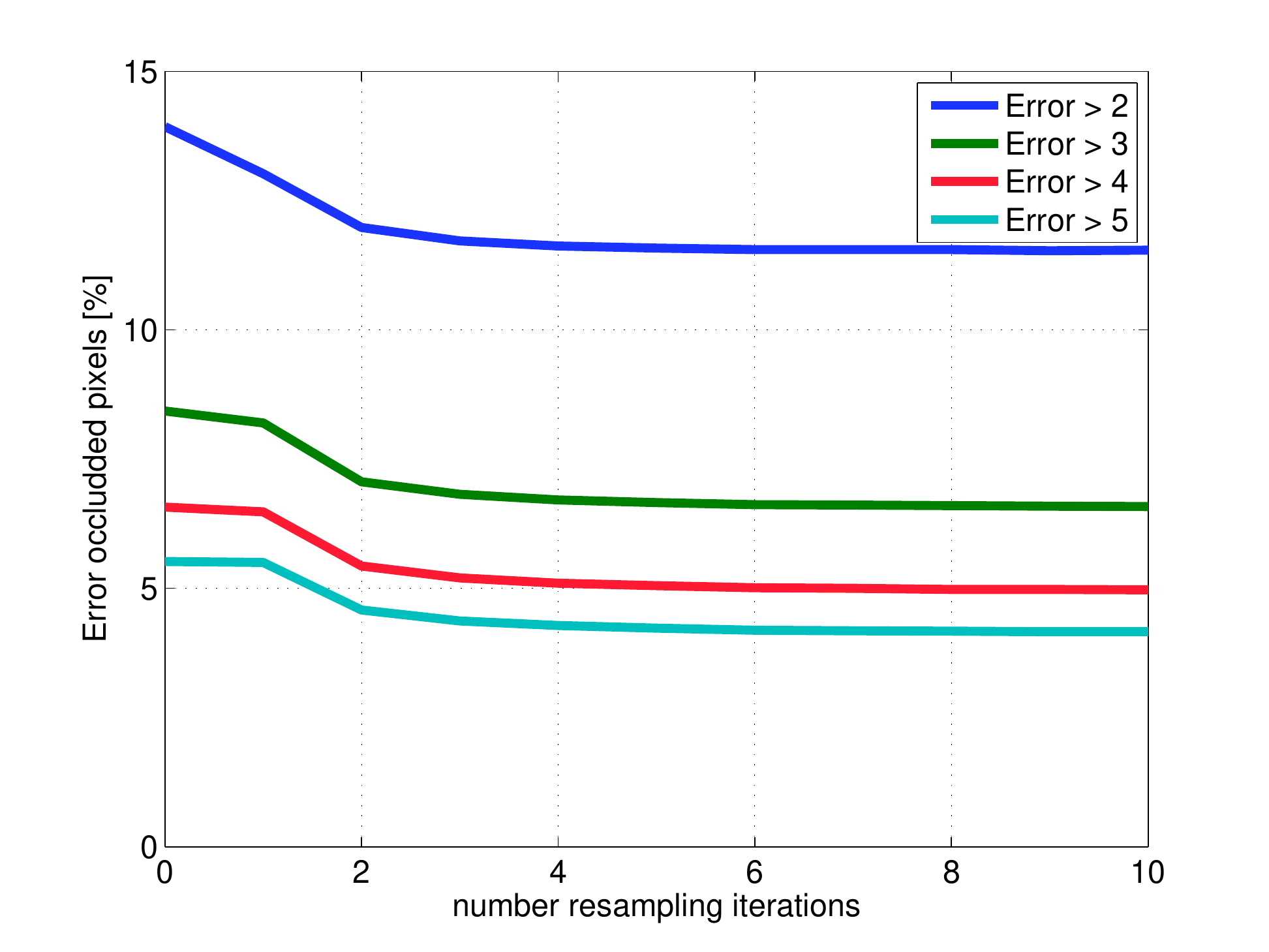}
\vspace{-0.4cm}
\caption{{\bf Importance of the number of re-sampling iterations:} on KITTI. }
\label{fig:kitti_re}
\end{center}
\end{figure}

\begin{table}[t]
\begin{center}
\begin{small}
\begin{tabular}{| c | c | c | c | c | c | c | c | c |}
\hline
Number of & \multicolumn{2}{|c|}{  $>$ 2 pixels} & \multicolumn{2}{|c|}{  $>$ 3 pixels}& \multicolumn{2}{|c|}{  $>$ 4 pixels}& \multicolumn{2}{|c|}{  $>$ 5 pixels}\\ \cline{2-9}
training images& {\bf Non-Occ} & {\bf Occ} & {\bf Non-Occ} & {\bf Occ}& {\bf Non-Occ} & {\bf Occ}& {\bf Non-Occ} & {\bf Occ}
\\ \hline
 1 & 9.14 \% & 11.49 \% & 4.58 \% & 6.55 \% & 3.23 \% & 4.92 \% & 2.61 \% & 4.09 \% \\ \hline
 5 & 9.02 \% & 11.58 \% & 4.46 \% & 6.66 \% & 3.13 \% & 5.06 \% & 2.50 \% & 4.24 \% \\ \hline
 10 & 9.03 \% & 11.58 \% & 4.47 \% & 6.66 \%  & 3.13 \% & 5.05 \% & 2.51 \% & 4.23 \% \\ \hline
 20 & 9.03 \% & 11.59 \% & 4.46 \% & 6.66 \% & 3.12 \% & 5.05 \% & 2.49 \% & 4.22 \% \\ \hline
\end{tabular}
\end{small}
\end{center}
\vspace{-0.3cm}
\caption{{\bf Training set size:} Estimation errors as a function of the training set size. Note that very few images are needed to learn good parameters.}
\label{tab:size}
\end{table}

We now describe the characteristics of the databases we evaluate our approach on. 
Our first dataset consists on high resolution images from the Middebury dataset \cite{Middlebury}, which have an average resolution of $1239.2\times1038.0$ pixels. We employ 5 images for training (i.e., {\it Books, Laundry, Moebius, Reindeer, Bowling2}) and 9 images for testing (i.e., {\it Cones, Teddy, Art, Aloe, Dolls, Baby3, Cloth3, Lampshade2, Rocks2}). 
We also evaluate our approach on the KITTI dataset \cite{Geiger12}, which  is the only real-world stereo dataset with accurate ground truth. 
It is composed of 194 training and 195 test high-resolution images ($1237.1 \times 374.1$ pixels) captured from an autonomous driving platform driving around in a urban environment. The ground truth is generated by means of a Velodyne sensor which is calibrated with the stereo pair. This results in  semi-dense ground truth covering approximately 30 \% of the pixels.  We employ 10 images for training, and utilize the remaining 184 images for validation purposes. 

\paragraph{{\bf Comparison with the state-of-the-art}:}
We begin our experimentation by comparing our approach with the state-of-the-art.
Table \ref{tab:kitti} depicts results of our approach and the baselines in terms of the two metrics for the KITTI dataset. 
Note that our approach significantly outperforms all the baselines in all settings (i.e., thresholds bigger than 2, 3, 4 and 5 pixels). 
Table \ref{tab:midd} depicts similar comparisons for high resolution Middlebury. Once more, our approach outperforms significantly  the baselines in all settings. Fig. \ref{fig:example} depicts an illustrative set of example results for the KITTI dataset. Note that despite the challenges that the images pose, our approach does a good job at estimating disparities. 
%
%


\paragraph{{\bf Segmentation strategy:}}
We next investigate how the segmentation strategy affects the stereo estimation. Towards this goal we evaluate the results of our approach when employing UCM segments \cite{j:Arbelaez11}, SLIC superpixels \cite{SLIC} or the intersection of both as input. Table \ref{tab:ucm_kitti} depicts results on the KITTI dataset. 
UCM performs very poorly as  the number of superpixels on average is very small, and some of the superpixels are very large. Therefore, a single 3D plane is a poor representation for the disparities in those large segments. 
SLIC performs quite well, but the intersection of SLIC and UCM superpixels outperforms the other strategies. This is also expected, as UCM respects the boundaries much better than SLIC. Note that  as shown in Table \ref{tab:ucm_midd} similar results are observed for the Middlebury dataset.

\begin{table}[t]
\begin{center}
\begin{small}
\begin{tabular}{ | c | c | c | c | c | c | c | c | c |}
\hline
 & \multicolumn{2}{|c|}{  $>$ 2 pixels} & \multicolumn{2}{|c|}{  $>$ 3 pixels} & \multicolumn{2}{|c|}{  $>$ 4 pixels} & \multicolumn{2}{|c|}{  $>$ 5 pixels} \\
 \cline{2-9}
 & {\bf N.-Occ} & {\bf Occ} & {\bf N.-Occ} & {\bf Occ} & {\bf N.-Occ} & {\bf Occ} & {\bf N.-Occ} & {\bf Occ} \\
 \hline
 Oracle 
 & 1.38\% & 1.70\% & 1.03\% & 1.27\% & 0.90\% & 1.10\% & 0.82\% & 0.99\% \\
 \hline
 Initial fit 
 & 10.69\% & 13.72\% & 6.12\% & 8.83\% & 4.59\% & 7.02\% & 3.76\% & 5.98\% \\
 \hline
 {\bf Ours} 
 & 9.64\% & 12.24\% & 5.14\% & 7.35\% & 3.70\% & 5.63\% & 2.97\% & 4.69\% \\
 \hline
\end{tabular}
\end{small}
\end{center}
\vspace{-0.3cm}
\caption{{\bf Oracle performance:} Oracle, our approach and initial fit on KITTI.}
\label{tab:oracle_kitti}
\end{table}

\begin{table}[t]
\begin{center}
\begin{small}
\begin{tabular}{| c | c | c | c | c | c | c | c | c | c | c |}
\hline
 & \multicolumn{2}{|c|}{  $>$ 1 pixel} & \multicolumn{2}{|c|}{  $>$ 2 pixels} & \multicolumn{2}{|c|}{  $>$ 3 pixels} & \multicolumn{2}{|c|}{  $>$ 4 pixels} & \multicolumn{2}{|c|}{  $>$ 5 pixels} \\
 \cline{2-11}
 & {\bf N.-Occ} & {\bf Occ} & {\bf N.-Occ} & {\bf Occ} & {\bf N.-Occ} & {\bf Occ} & {\bf N.-Occ} & {\bf Occ} & {\bf N.-Occ} & {\bf Occ} \\
 \hline
 Oracle & 2.0\% & 6.2\% & 1.4\% & 5.5\% & 1.3\% & 5.3\% & 1.2\% & 5.2\% & 1.1\% & 5.1\% \\
 \hline
 Initial fit & 6.1\% & 21.8\% & 4.2\% & 19.3\% & 3.5\% & 18.3\% & 3.2\% & 17.7\% & 3.0\% & 17.3\% \\
 \hline
 {\bf Ours} & 5.1\% & 17.5\% & 3.3\% & 14.3\% & 2.8\% & 13.0\% & 2.5\% & 12.2\% & 2.3\% & 11.6\% \\
 \hline
\end{tabular}
\end{small}
\end{center}
\vspace{-0.3cm}
\caption{{\bf Oracle performance:} Oracle, our approach and initial fit on Middlebury.}
\label{tab:oracle_midd}
\end{table}

\paragraph{{\bf Number of superpixels:}}
We next investigate how well our approach scales with the number of superpixels in terms of computatinal complexity as well as accuracy. Fig. \ref{fig:time} shows results for the KITTI dataset when varying the number of superpixels. Our approach reduces performance gracefully when reducing the amount of superpixels. Note that  inference scales linearly with the number of superpixels, taking on average $5.5$ minutes per high resolution image when employing 1200 superpixels and $2.5$ minutes when using 300.

\paragraph{{\bf Number of re-sampling iterations:}} 
We evaluate the effects of varying the number of resampling iterations on the performance of our approach. As shown in Fig. \ref{fig:kitti_re}, our approach  converges to a good local optima after only 2 resampling iterations. This reduces the inference cost from 5.5 minutes per high-resolution image for 5 iterations to 2.2 minutes for 2 iterations.

\paragraph{{\bf Training set size:}} We evaluate the effect of increasing the training set size in Table \ref{tab:size}. Even when training with a single image we outperform all baselines.

\paragraph{{\bf Oracle performance:}}
We next evaluate the best performance that our model can achieve, by fitting the model to the ground truth disparities. This shows an upper-bound on the performance that our method could ever achieve if we were able to learn an energy that has its MAP at the ground truth, and if we were able to solve the NP-hard inference problem. 
Tables \ref{tab:oracle_kitti} and \ref{tab:oracle_midd} depict the oracle performance  in terms of both the occluded an non-occluded pixels for both datasets. 
Note that as KITTI does not release the test ground truth, we compute this values using 10 images for training and the rest of the training set for testing. 
We also report   performance of our initialization which is computed by fitting a local plane to the results of semi-global block matching \cite{Hirschmueller08}. 
Note that the oracle can achieve great performance, showing that  the errors due to the 3D slanted plane discretization are negligible. 


\paragraph{{\bf Robustness to noise:}} We investigate the robustness of our approach to noise by building a synthetic dataset, which is composed of 10 images for training and 90 images for test.  Each Image has a resolution of $320\times240$ pixels and contains several planes. The average number of superpixels is 108.0. We create ${\cal D}(\bp)$ by sampling 3 to 5 points at random on the boundaries and generating disparities by corrupting the ground truth with Gaussian noise of varying standard deviation. 
Table \ref{tab:synth} shows RMS errors for disparity as well as  percentage of boundary variables wrongly estimated.


\begin{table}[t]
\begin{center}
\begin{small}
\begin{tabular}{|c|c|c|}
\hline
Noise & RMS (pixels) &  boundary error \\
\hline
0 & 0.44 & 0.3 \% \\
\hline
1 & 0.80 & 0.6 \% \\
\hline
2 & 1.37 & 1.9 \% \\
\hline
3 & 2.24 & 5.3 \% \\
\hline
5 & 4.40 & 8.9 \% \\
\hline
\end{tabular}
\end{small}
\end{center}
\vspace{-0.3cm}
\caption{{\bf Robustness to noise}: RMS as well as boundary error as a function of noise.}
\label{tab:synth}
\end{table}

\paragraph{{\bf Family of structure prediction problems:}}

We evaluate the performance of our learning algorithm as a function of its parameter $\epsilon$ which ranges from CRFs for $\epsilon=1$ to structural SVMs for $\epsilon=0$. Fig. \ref{tab:epsilon} depicts performance on the KITTI dataset. Our approach results in state-of-the-art performance for all settings. 

\begin{table}[t]
\begin{center}
\begin{small}
\begin{tabular}{| c | c | c |c | c | c | c | c | c |}
\hline
& \multicolumn{2}{|c|}{  $>$ 2 pixels} & \multicolumn{2}{|c|}{  $>$ 3 pixels}& \multicolumn{2}{|c|}{  $>$ 4 pixels}& \multicolumn{2}{|c|}{  $>$ 5 pixels}\\ \hline
& {\bf Non-Occ} & {\bf Occ} & {\bf Non-Occ} & {\bf Occ}& {\bf Non-Occ} & {\bf Occ}& {\bf Non-Occ} & {\bf Occ}
\\ \hline
$\epsilon = 0.0$ & 9.16 \% & 11.84 \% & 4.64 \% & 6.96 \% & 3.25 \% & 5.30 \% & 2.58 \% & 4.42 \% \\ \hline
$\epsilon = 0.5$ & 9.15 \% & 11.59 \% & 4.59 \% & 6.65 \% & 3.24 \% & 5.02 \% & 2.59 \% & 4.17 \% \\ \hline
$\epsilon = 1.0$ & 9.03 \% & 11.58 \% & 4.47 \% & 6.66 \% & 3.13 \% & 5.05 \% & 2.51 \% & 4.23 \% \\ \hline
\end{tabular}
\end{small}
\end{center}
\vspace{-0.2cm}
\caption{{\bf Family of structure prediction problems:} for the KITTI dataset. Note that $\epsilon=0$ is structural SVM and $\epsilon=1$ is CRF.}
\label{tab:epsilon}
\end{table}

\paragraph{{\bf Task Loss:} } We evaluate the importance of incorporating different tasks loss in the learning framework of \cite{Hazan10}. In particular, we employ RMS as well as the same loss that we employ for evaluation. Note that the loss has little effect.



\begin{table}[t]
\begin{center}
\begin{small}
\begin{tabular}{| c | c | c |c | c | c | c | c | c |}
\hline
& \multicolumn{2}{|c|}{  $>$ 2 pixels} & \multicolumn{2}{|c|}{  $>$ 3 pixels}& \multicolumn{2}{|c|}{  $>$ 4 pixels}& \multicolumn{2}{|c|}{  $>$ 5 pixels}\\ \hline
& {\bf Non-Occ} & {\bf Occ} & {\bf Non-Occ} & {\bf Occ}& {\bf Non-Occ} & {\bf Occ}& {\bf Non-Occ} & {\bf Occ}
\\ \hline 
$\ell_2$ loss & 9.03 \% & 11.58 \% & 4.47 \% & 6.66 \%  & 3.13 \% & 5.05 \% & 2.51 \% & 4.23 \% \\ \hline
 $>$2 pixels & 9.03 \% & 11.59 \% & 4.46 \% & 6.65 \% & 3.13 \% & 5.05 \% & 2.50 \% & 4.24 \% \\ \hline
$>$3 pixels & 9.05 \% & 11.58 \% & 4.47 \% & 6.64 \% & 3.13 \% & 5.03 \% & 2.49 \% & 4.20 \% \\ \hline
 $>$4 pixels & 9.02 \% & 11.58 \% & 4.47 \% & 6.67 \% & 3.13 \% & 5.06 \% & 2.50 \& & 4.23 \% \\ \hline
 $>$5 pixels & 9.05 \% & 11.60 \% & 4.48 \% & 6.66 \% & 3.13 \% & 5.04 \% & 2.50 \% & 4.22 \% \\ \hline
\end{tabular}
\end{small}
\end{center}
\caption{{\bf Loss functions}: Performance changes when employing different loss functions for learning  in the KITTI dataset}
\label{tab:loss}
\end{table}

\section{Conclusion}

We have presented a novel stereo slanted-plane MRF model that reasons jointly about occlusion boundaries as well as depth. We have formulated the problem as inference in a hybrid MRF composed of both continuous (i.e., slanted 3D planes) and discrete (i.e., occlusion boundaries) random variables, which we have tackled using particle convex belief propagation. 
We have demonstrated the effectiveness of our approach on high resolution imagery from Middlebury as well as the more challenging KITTI dataset. 
In the future we plan to investigate alternative inference algorithms as well as other segmentation potentials.

\bibliographystyle{splncs}
\bibliography{biblio}

\end{document}